\documentclass[10pt,letterpaper]{article}

\usepackage{cogsci}

 \cogscifinalcopy %

\usepackage[
  style=apa,
  natbib=true,
  annotation=false,
]{biblatex}
\RequirePackage{xcolor}
\definecolor{blue}{HTML}{1F77B4}
\definecolor{orange}{HTML}{FF7F0E}
\definecolor{green}{HTML}{2CA02C}
\definecolor{red}{HTML}{D62728}
\usepackage{graphicx}
\usepackage{booktabs}
\usepackage{multirow}
\usepackage{bm}
\usepackage{framed}
\renewenvironment{leftbar}{%
  
  \MakeFramed {\advance\hsize-\width \FrameRestore}}
{\endMakeFramed}

\title{Are they human? Detecting large language models\\by probing human memory constraints}

\author[1]{\mbox{Simon Schug}} %
\author[1,2]{\mbox{Brenden M. Lake}}
\affil[1]{Department of Computer Science, Princeton University}
\affil[2]{Department of Psychology, Princeton University}

\begin{document}
\renewcommand{\thefootnote}{\fnsymbol{footnote}}
\maketitle
\renewcommand{\thefootnote}{\arabic{footnote}}
\setcounter{footnote}{0}

\begin{abstract}

The validity of online behavioral research relies on study participants being human rather than machine.
In the past, it was possible to detect machines by posing simple challenges that were easily solved by humans but not by machines.
General-purpose agents based on large language models (LLMs) can now solve many of these challenges, threatening the validity of online behavioral research.
Here we explore the idea of detecting humanness by using tasks that machines can solve too well to be human.
Specifically, we probe for the existence of an established human cognitive constraint: limited working memory capacity.
We show that cognitive modeling on a standard serial recall task can be used to distinguish online participants from LLMs even when the latter are specifically instructed to mimic human working memory constraints.
Our results demonstrate that it is viable to use well-established cognitive phenomena to distinguish LLMs from humans.

\end{abstract}

\section{Introduction}

Ensuring that online behavioral experiments are performed by humans rather than machines is an obvious requirement for their validity.
Traditionally, machines could be detected by posing challenges like simple logic puzzles, attention checks or catch trials \citep{abbey_attention_2017, douglas_data_2023}.
Foundation models, and large language models (LLMs) in particular can now perform well on many of these tasks and autonomously complete online surveys in ways that are difficult to distinguish from humans \citep{rilla_recognising_2025, westwood_potential_2025}.
Therefore, finding new tasks for detecting non-human participants -- that are suitable for online experiments and can be easily solved by humans but not by machines -- is becoming increasingly difficult.
\begin{leftbar}
\noindent Rather than relying on machine limitations, can we identify humans by probing for the unique constraints of human cognition?
\end{leftbar}

Cognitive science has a rich tradition of characterizing the behavioral signatures and cognitive constraints of humans.
LLMs, despite being trained on large quantities of human-generated data, may not inherit certain human cognitive constraints, especially if these constraints are misaligned with the training objective and are absent in the model architecture itself.
There are many candidates for established human cognitive constraints that LLMs might not inherit.
For example, humans get tired, perceive optical illusions \citep{ullman_illusion-illusion_2024}, and can only store few items in their working memory \citep[e.g.][]{miller_magical_1956, cowan_magical_2001}.
Here, we focus on the latter and investigate whether it is possible to distinguish online participants from LLMs on a classic serial probed recall working memory task \citep{murdock_serial_1968}.

If LLMs were to display similar behavioral signatures as humans on this task it could either be because they are genuinely subject to the same cognitive constraints or because they have learned to imitate human response patterns, especially when explicitly instructed to imitate them.
A priori, the particular cognitive constraints humans show in serial recall working memory tasks are unlikely to be shared by LLMs.
Current LLMs are based on the transformer architecture and employ multi-head attention which allows for direct access to information in the context window which can span hundreds of thousands to millions of tokens \citep{peng_yarn_2024, team_gemini_2024}.
Furthermore, the ability for long contextual recall is beneficial both for the training objective of next token prediction during pre-training or reward maximization during post-training.
However, it is likely that many LLMs are trained on human behavioral data from psychological experiments given the public availability of such datasets \citep{binz_foundation_2025}.
It stands to question whether this suffices to produce humanlike data, given that the quantitative effects strongly depend on the specific experimental instantiations of working memory tasks \citep{oberauer_benchmarks_2018}.
In the following we will explore whether LLMs exhibit systematically different error patterns on working memory probes -- even when specifically instructed to behave humanlike -- and can be distinguished from online participants on this basis.

\section{Methods}

\begin{figure*}[t]
  \centering
  \includegraphics[width=\textwidth]{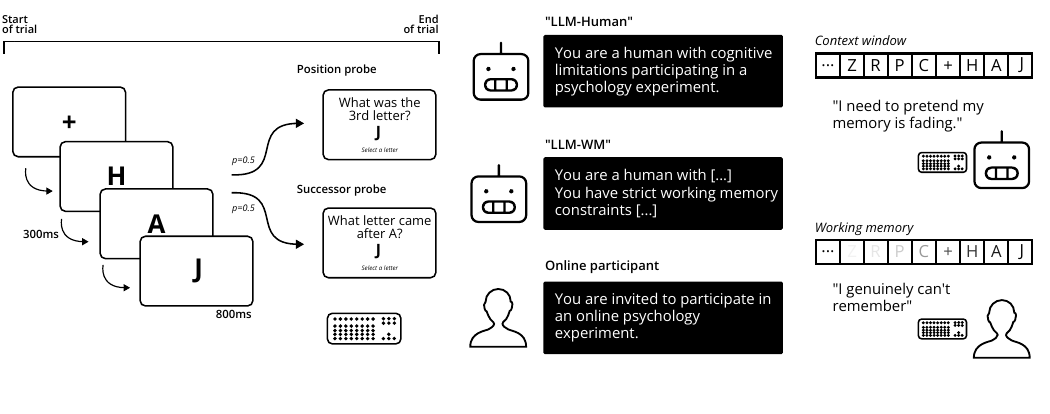}
\vspace{-1cm}
  \caption{\textbf{Overview}. \textbf{\textit{Left}} Probed recall working memory paradigm with two probe type conditions chosen uniformly at random within each trial. \textbf{\textit{Center}} We compare participants recruited online to a sample of large language models (LLMs) with varying system prompts and backbone models on the probed recall working memory task. \textbf{\textit{Right}} In contrast to humans, transformer-based LLMs have perfect contextual recall through the attention mechanism and need to emulate serial recall effects.}
  \label{fig:graphical-abstract}
\end{figure*}

\begin{figure*}[t]
  \centering
  \includegraphics[width=0.25\textwidth]{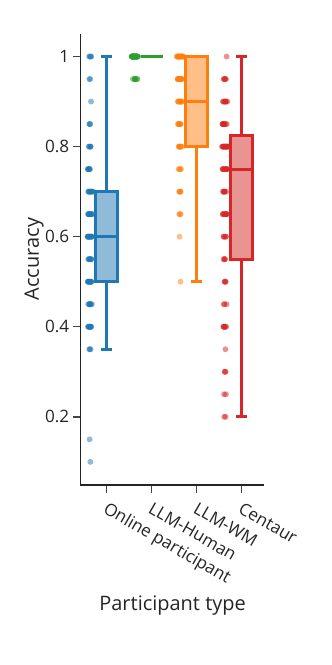}
  \includegraphics[width=0.37\textwidth]{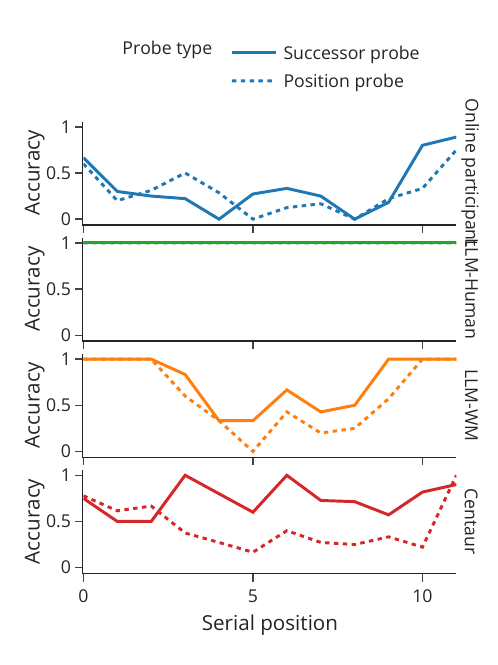}
  \includegraphics[width=0.37\textwidth]{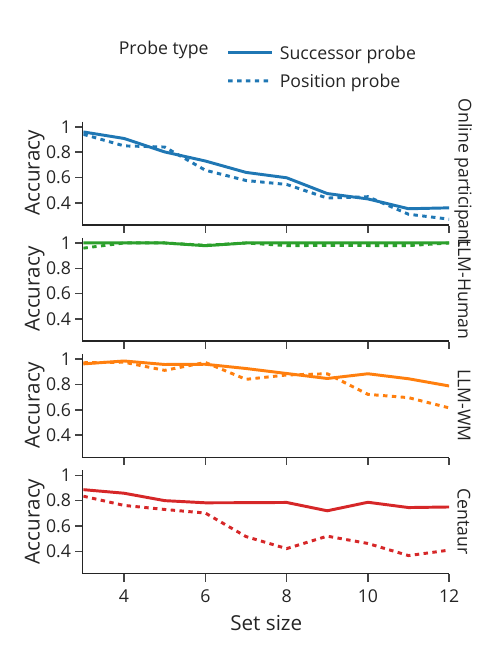}
\vspace{-0.75cm}
  \caption{
  \textbf{Working memory performance comparison on probed recall task}. \textbf{\textit{Left}} Box plot showing average accuracies by participant type. Each box spans from the first quartile to the third quartile with the middle line denoting the median and whiskers reach 1.5 times the interquartile range from the box. Accuracy across participants varies significantly between online participants and LLMs. Despite being instructed to behave human-like, LLMs have near perfect accuracy (LLM-Human). Average accuracy only decreases, when additionally instructed to display specific working memory limitations (LLM-WM) or finetuned on data from psychology experiments (Centaur). \textbf{\textit{Center}} Average accuracy as a function of serial position index on trials with a set size of 12. Online participants show established serial position effects, remembering items early on (primacy) and at the end of the list (recency) more readily. LLMs show no such effect when instructed to behave human-like (LLM-Human) but qualitatively reproduce these effects when specifically instructed to do so (LLM-WM) or after finetuning (Centaur). \textbf{\textit{Right}} Average accuracy for each set size. Online participants show a clear set-size effect on accuracy, with performance decreasing due to increasing memory load as the set size grows. Again, LLMs qualitatively reproduce this effect when specifically instructed to do so (LLM-WM) or finetuned (Centaur) but not when instructed to behave human-like (LLM-Human).}
  \label{fig:working-memory-performance}
\end{figure*}

\subsection{Paradigm}

We employ a standard probed recall working memory paradigm \citep{murdock_serial_1968, oberauer_benchmarks_2018}.
In each trial, participants are sequentially presented with a list of letters and have to recall a particular letter based either on a position probe (e.g., "What was the 3rd letter?") or on a successor probe (e.g., "What letter came after X?"), as illustrated in Figure~\ref{fig:graphical-abstract}.
Each letter is presented for $800$ms and the interstimulus interval is set to $300$ms.
All participants complete $T=20$ trials in total with set sizes varying between 3 to 12 letters in random order but balanced within participants to contain each set size exactly twice.
The serial position of the probe is chosen uniformly at random.
Prior to performing the trials, participants are instructed on how the task works, followed by a brief instruction comprehension quiz and 4 practice trials.
The full experiment is implemented in the smile library \citep{gureckis_smile_2022} and hosted on a web server allowing online participants to access it through their personal laptop or desktop computer in a web browser.

\subsection{LLM interface}
The technical expertise required for deploying LLM-based agents to participate in online experiments lowers as browser automation tools for LLMs mature and become more widely available.
Indeed off the shelf LLM-based browser automation tools such as "Gemini for Chrome" \citep{google_gemini_2026} are able to autonomously navigate and complete our online experiment.
However, current browser automation tools primarily rely on static snapshots of a website.
Such snapshots fail to capture the rapidly presented list of letters during our working memory task.
As a result, LLM-based agents can struggle on our task solely due to the imperfect interface.
While this issue can be easily mitigated, e.g. by instructing the model to directly access the application state, we would like to separate errors due to specific problems in the interface from the genuine error patterns of an LLM agent on our task.
Therefore, we will evaluate LLM participants on a streamlined textual interface that presents all the information an online participant sees but removes any markup and styling.
Each LLM participant is then run through the whole experiment, including instructions and instruction quiz, in one continuous conversation and expected to return structured output for each choice that needs to be made throughout the experiment.

\subsection{Participants}

We collect data from $100$ online participants recruited through Prolific \citep{palan_prolificacsubject_2018}.
Participants are required to reside in the United States of America, to be at least 18 years old and to be fluent English speakers.
Each participant was paid \$4 and it took participants around 12 minutes on average to complete the experiment. 
For the LLM participants we will be conducting two simulations with different system prompts we will describe below and access current LLMs from different model providers through their respective APIs using their respective default configurations for sampling parameters and thinking/reasoning budgets.
The specific providers and models are listed in Table~\ref{tab:llm-wm-detection} using model identifiers as defined in the llm python library \citep{willison_llm_2023} as of 2026-02-02.
In total we collect 5 independent seeds for each of the 11 models resulting in 55 distinct LLM participants per simulation or a total of 205 participants -- human and machine.\footnote{
A bug in the llm python library \citep{willison_llm_2023} prevented us from evaluating \texttt{claude-opus-4.5} and \texttt{claude-haiku} with structured outputs.
We report results for the slightly older \texttt{claude-opus-4.1} model.
}

\subsection{Measures against uncontrolled LLM participants}

While our recruitment platform Prolific implements additional measures to ensure that participants are real humans, we cannot exclude the possibility that some of our participants are nevertheless themselves using LLM-based agents.
To alleviate this issue, we implement additional detection measures.
We include an AI self-report, directly asking participants whether they are themselves AIs, and randomly include one of three questions designed to be hard to answer by a human but easy for current LLMs that participants were allowed to skip.
This includes two questions in low resource languages: Māori, a language spoken by indigenous Polynesian people of mainland New Zealand, with less than 150\,000 people reporting conversational level proficiency in a census from 2013 \citep{stats_nz_tatauranga_aotearoa_2013_2013} and Võro, a language spoken in South Estonia with roughly 75\,000 speakers \citep{statistikaamet_eesti_2011}.
The third question involves recalling a hexadecimal number after the experiment -- that participants needed to enter to proceed from the task instructions but were not instructed to remember -- and converting it to a decimal number.
Neither the AI self-report nor the questions designed to detect naive LLMs revealed the presence of LLMs in our sample of online participants.
In our own simulations, these measures were able to successfully detect naive LLMs but could be circumvented by explicitly instructing models to avoid specific actions -- for instance, to only speak English.

\section{Simulation 1:\\Detecting LLMs instructed to be human}

In this first part, we will focus on comparing online participants to LLMs that are instructed to be human with the following system prompt:

\begin{leftbar}
{
\color{green}
\ttfamily
\noindent You are a human with cognitive limitations participating in a psychology experiment.
}
\end{leftbar}
\noindent This prompt is deliberately simple to reflect an LLM-based agent participant that attempts to be generally useful across different kinds of surveys and online experiments.
We refer to LLM participants that use this prompt as \textcolor{green}{LLM-Human}.

\subsection{Results}
Figure~\ref{fig:working-memory-performance} shows participants' average accuracies across all trials.
LLM participants with the LLM-Human system prompt achieve almost perfect accuracy across trials with a median accuracy of $1.00$, showing no serial position or set size effects regardless of the probe type.
As a result, it is straightforward to detect and exclude all LLM-Human participants with a simple threshold on the average task accuracy of 0.95.
Applying the same criterion to the online participants leads to the exclusion of 6 of the 100 online participants which had (close to) perfect performance despite a median accuracy of $0.6$ across online participants.

\subsection{Discussion}

Excluding participants based on unusually high accuracies is an established exclusion criterion for online working memory experiments in which it is difficult to ensure that participants are not cheating e.g. by using external aids rather than relying on short-term memory \citep{young_comparison_2025}.
In this case, it is also a simple and effective measure to remove all LLM participants.

While the fact that the transformer-based LLMs we evaluated here \textit{can} solve the probed recall task close to perfectly might come at no surprise, one might have expected the instruction to behave humanlike to have a stronger effect on performance. 
It is therefore reasonable to ask, if changing the system prompt to be more explicit in what way the LLM is supposed to behave humanlike, would lead to LLM participants that are no longer distinguishable from online participants.
We will consider this scenario in the next section.

\begin{figure*}[t]
  \centering
  \includegraphics[width=0.25\textwidth]{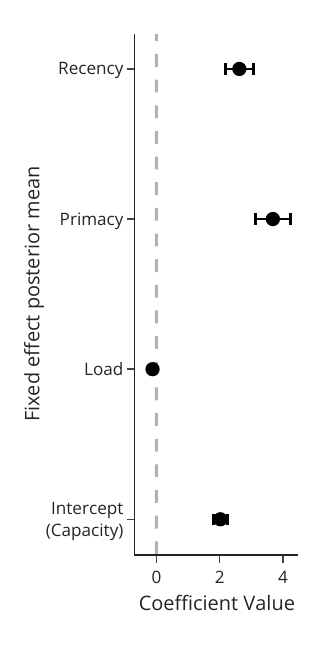}
  \includegraphics[width=0.37\textwidth]{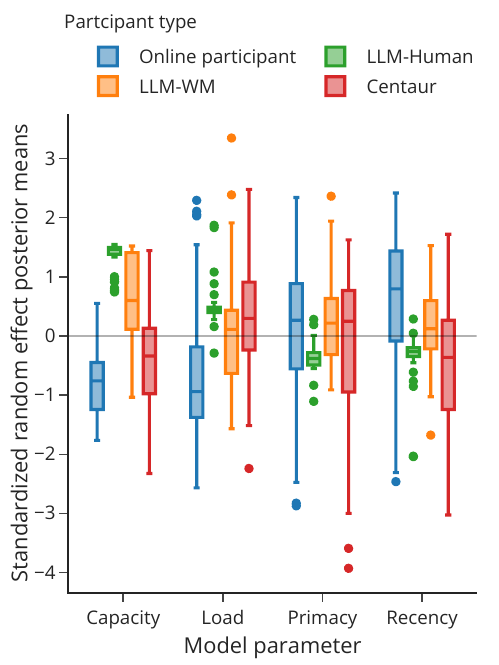}
  \includegraphics[width=0.37\textwidth]{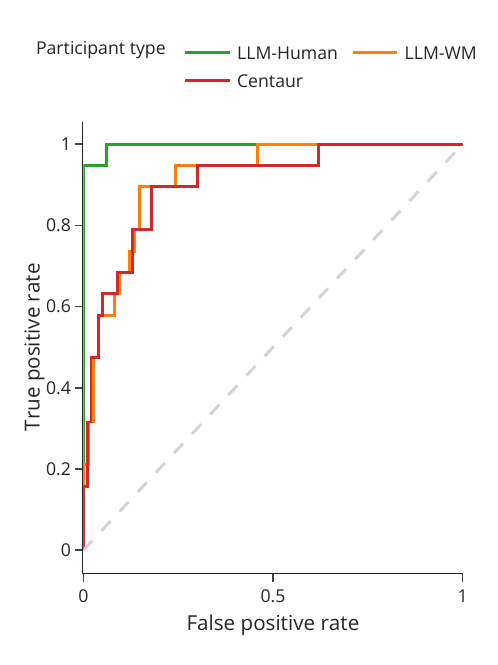}
  \vspace{-0.75cm}
  \caption{\textbf{Working memory profile comparison and anomaly detection}. We use a hierarchical Bayesian logistic regression model to capture the probability of giving a correct response in the probed recall task based on established working memory phenomena. \textbf{\textit{Left}} Jointly fitting this model on both online and LLM participants, we find strong fixed-effects for recency, primacy and capacity. Whiskers denote the 94\% highest density interval (HDI). \textbf{\textit{Center}} The joint model reveals that participant-level working memory profiles captured by their random effects differ notably between participant types. \textbf{\textit{Right}} We perform anomaly detection by fitting a second model on a subset of online participants and scoring held-out online participants and LLM participants with the model using log predictive densities. The resulting ROC reveals that LLM participants (negative class) can be separated from online participants (positive class) even when designed to mimic working memory.}
  \label{fig:wm-profiles}
\end{figure*}

\section{Simulation 2: Detecting LLMs explicitly designed to mimic working memory}

We now consider an adversarial setting where LLMs are explicitly designed to mimic the specific working memory constraints that we use to distinguish online participants from LLM participants to evaluate the robustness of the approach.
In particular, we will use models with an extended system prompt with explicit working memory instructions and a model finetuned on data from human psychological experiments.
We then present a strategy to detect such LLMs using cognitive anomaly detection.

\subsection{LLMs instructed to mimic working memory}
We instruct LLMs to mimic working memory constraints by extending the system prompt from before as follows:

\begin{leftbar}
{
\footnotesize
\color{orange}
\ttfamily
\noindent You have strict working memory limitations -- you can only hold a limited number of items in your short-term memory.
When presented with a long list of items without rehearsal opportunities, you will experience memory decay, particularly for items in the middle of the list.
\begin{enumerate}
    \item You must process the items sequentially as they appear
    \item You must forget items based on serial position effects - remembering beginning items (primacy) and recent items (recency) better than middle items
    \item You must introduce errors in recall according to these serial position effects.
\end{enumerate}
}
\end{leftbar}

\noindent We refer to LLM participants that use this prompt as \textcolor{orange}{LLM-WM}.
As can be observed in Figure~\ref{fig:working-memory-performance} using this system prompt reduces average accuracy on the probed recall task to a median of 0.90 with a far larger spread across LLM-WM participants.
While this is still notably higher than our online participants, it is no longer possible to exclude all LLM-WM participants merely through a simple accuracy threshold without also removing a significant portion of online participants.
As instructed, LLM participants with the LLM-WM system prompt qualitatively display both primacy and recency serial position effects and a set size effect.

\subsection{LLMs finetuned on data from human experiments}

While general purpose LLMs might not inherit human cognitive constraints, models specifically trained on human psychological data are being developed exactly with this purpose in mind: to emulate human response patterns.
To evaluate to what extent such models can already successfully generate humanlike data on our working memory task we consider the Centaur model \citep{binz_foundation_2025}:

\begin{leftbar}
{
\footnotesize
\color{red}
\ttfamily
\noindent Centaur is an LLM that has been finetuned on a large corpus of trial-by-trial natural language transcripts from human psychological experiments. 
}
\end{leftbar}

\noindent 

The left panel of Figure~\ref{fig:working-memory-performance} shows that the overall task accuracy of Centaur participants resembles our sample of online participants even more than LLMs instructed via their system prompt to specifically emulate working memory constraints.
For the position-based probe, both the serial position effects in the center panel and the set size effect in the right panel look deceivingly similar to the online participants, whereas for the successor probe there remain visible differences.

\subsection{Cognitive anomaly detection}

Despite the qualitative similarities on the aggregate task statistics, it might be possible to distinguish participants based on their specific, quantitative working memory profiles.
In the following, we will therefore treat the problem of LLM detection as a cognitive anomaly detection problem using a hierarchical Bayesian working memory model.
We use this model in two ways:
\begin{enumerate}
  \item \textbf{Descriptive analysis model}: We fit the parameters of the model on all participants -- online participants and LLM participants -- to obtain a quantitative description of how working memory profiles differ between the groups.
  \item \textbf{Anomaly detection model}: We fit the parameters of the model solely on a subset of online participants, and use the resulting model to perform anomaly detection by scoring the plausibility of unseen participants' data.
\end{enumerate}

Specifically, we use a hierarchical Bayesian logistic regression model to capture established effects in the working memory literature \citep{oberauer_benchmarks_2018}: variations in short-term memory capacity, a set size effect of increasing memory load, as well as serial position effects where items early and late in the sequence are remembered more readily.
\newpage
Concretely, we model the probability of a correct response for participant $i$ on trial $t$ as
\begin{align*}
y_{i,t} &\sim \mathrm{Bernoulli}(p_{i,t}), \\
\text{logit}(p_{i,t}) &= \beta_{i}^{\text{capacity}} \\
&+ \beta_{i}^{\text{load}} x_{i,t}^{\text{load}} \\
&+ \beta_{i}^{\text{primacy}} x_{i,t}^{\text{primacy}} \\
&+ \beta_{i}^{\text{recency}} x_{i,t}^{\text{recency}},
\end{align*}
where $x_{i,t}^{\text{load}}$, $x_{i,t}^{\text{primacy}}$, and $x_{i,t}^{\text{recency}}$ denote mean-centered set size, primacy and recency respectively.
The primacy and recency covariates, $x_{i,t}^{\text{primacy}}$ and $x_{i,t}^{\text{recency}}$, are defined as the inverse serial position of the probe from the beginning and from the end of the sequence, respectively.

Participant-specific coefficients decompose into population and individual components,
\begin{align*}
\beta_i^k &= \mu^k + \gamma_i^k, \quad k \in \{\text{capacity}, \text{load}, \text{primacy}, \text{recency}\} \\
\bm{\mu} &= (\mu^{\text{capacity}}, \mu^{\text{load}}, \mu^{\text{primacy}}, \mu^{\text{recency}})^\top,\\
\bm{\gamma}_i &= (\gamma_i^{\text{capacity}}, \gamma_i^{\text{load}}, \gamma_i^{\text{primacy}}, \gamma_i^{\text{recency}})^\top, \\
\bm{\sigma} &= (\sigma^{\text{capacity}}, \sigma^{\text{load}}, \sigma^{\text{primacy}}, \sigma^{\text{recency}})^\top,
\end{align*}
\noindent where $\mu^k \in \mathbb{R}$ are population-level, fixed effects, $\gamma_i^k \in \mathbb{R}$ are participant-level, random effects and $\sigma^k \in \mathbb{R}_{+}$ are mutually independent, population-level standard deviations of the random effects.
We use the weakly informative default priors of the PyMC-based Bambi package \citep{abril-pla_pymc_2023, capretto_bambi_2022} with independent, empirically scaled priors for fixed and random effects.

In addition to a descriptive analysis of all participants working memory profiles, we can use this model to detect anomalous participants by training it on a set of participants that are known to be normal and evaluating the likelihood of observing data from a new participant under the model.
Specifically, given a training set, $\mathcal{D}_{\text{train}} = \left \{ \left ( (\bm{x}_{i,t}, y_{i,t}) \right )_{t=1}^T \right \}_{i \in \mathcal{I}_{\text{train}}}$, we score data from an unseen participant, $\left ( (\bm{x}_{*,t}, y_{*,t}) \right )_{t=1}^T$, by computing the mean \textit{log pointwise predictive density},
\begin{align*}
\frac{1}{T} \sum_{t=1}^T \log \iint p(y_{*,t} \mid \bm{x}_{*,t}, \bm{\mu}, \bm{\sigma}) p(\bm{\mu}, \bm{\sigma} \mid \mathcal{D}_{\text{train}}) \, d\bm{\mu} \, d\bm{\sigma},
\end{align*}
where $p(y_{*,t} \mid \bm{x}_{*,t}, \bm{\mu}, \bm{\sigma})$ denotes the marginal likelihood for a new participant after integrating out their unobserved random effects,
\begin{align*}
p(y_{*,t} \mid \bm{x}_{*,t}, \bm{\mu}, \bm{\sigma}) = \int p(y_{*,t} \mid \bm{x}_{*,t}, \bm{\mu}, \bm{\gamma}_*) p(\bm{\gamma}_* \mid \bm{\sigma}) \, d\bm{\gamma}_*.
\end{align*}
Anomalous participants can then be detected by comparing the scores of new participants to the scores of a held-out set of normal participants.
While the joint predictive density is arguably a more powerful score, we find that the pointwise predictive density performs better in practice.
We expand on this observation in the discussion.

\begin{table}[t]
\centering
\begin{tabular}{llr}
\toprule
\textbf{Provider} & \textbf{Model identifier} & \textbf{AUROC} \\
\midrule
Anthropic & \texttt{claude-opus-4.1} & 1.0000 \\
Anthropic & \texttt{claude-sonnet-4.5} & 1.0000 \\ \midrule
Google & \texttt{gemini-2.5-flash} & 0.8737 \\
Google & \texttt{gemini-2.5-flash-lite} & 1.0000 \\
Google & \texttt{gemini-2.5-pro} & 0.8263 \\
Google & \texttt{gemini-3-flash-preview} & 0.7632 \\
Google & \texttt{gemini-3-pro-preview} & 0.9474 \\ \midrule
OpenAI & \texttt{gpt-5-mini} & 0.9263 \\
OpenAI & \texttt{gpt-5-nano} & 1.0000 \\
OpenAI & \texttt{gpt-5.2} & 0.9895 \\ \midrule \midrule
Hugging Face & \texttt{llama-3.1-centaur-70B} & 0.9026 \\
\bottomrule
\end{tabular}
\caption{Detection AUROC for LLM-WM participants with an explicit working memory system prompt and Centaur. Detection difficulty varies between different model backbones.}
\label{tab:llm-wm-detection}
\end{table}

\subsection{Results}

We perform posterior inference of the hierarchical Bayesian working memory model described above using the No-U-Turn Sampler \citep{hoffman_no-u-turn_2014} on four independent chains, warming-up for 2000 steps and drawing 2000 samples per chain from the posterior.
We begin by fitting data from all participants -- online and LLM-based -- in a single \textbf{descriptive analysis model} to quantify and compare participants' working memory profiles.
The left and center panel of Figure~\ref{fig:wm-profiles} show the resulting posterior means of the fixed and random effects.
The population-level fixed effects show strong serial position effects for both primacy and recency.
Items early in the list and late in the list are more likely to be remembered ($\mu^{\text{primacy}}= 3.68, 94\% \text{ HDI } [3.13, 4.22]$, $\mu^{\text{recency}}= 2.62, 94\% \text{ HDI } [2.19, 3.06]$) and increased memory load due to larger set sizes leads to decreased recall accuracy ($\mu^{\text{load}}= -0.12, 94\% \text{ HDI } [-0.16, -0.083 ]$).
The distribution over participant-level random effects grouped by participant type in the center panel of Figure~\ref{fig:wm-profiles} shows that these effects differ between online participants and the different LLM participants.
Online participants typically have a lower baseline memory capacity (intercept) than LLM participants and a stronger negative load effect.
However, there is substantial variation in the serial position effects of online participants and the distributions between participant types overlap as a result.

Next, we attempt to discriminate online participants from LLM participants without assuming access to ground truth labels for all participants in an \textbf{anomaly detection model}.
We only consider hard trials with set sizes of 9 to 12 items in this part since differences are most pronounced on hard trials.
To this end, we designate 80\% of the online participants' data as the training data to define our normative group. 
We fit the parameters of the working memory model on the training data and compute the mean log pointwise predictive density for each held-out online participant and all of the LLM participants.
The right panel of Figure~\ref{fig:wm-profiles} shows the resulting ROC for separating LLM participants (negatives) from online participants (positives) grouped by participant type based on the log pointwise predictive density scores.
Our anomaly detection model can almost perfectly detect LLM-Human participants, consistent with their abnormally high average task accuracy.
Detection of LLM-WM and Centaur participants is comparably more difficult, requiring a trade-off between erroneously including LLM-WM and Centaur participants (false positives), and mistakenly excluding online participants (false negatives).
For example, if we want to ensure that not more than 10\% of online participants are excluded, we would have to erroneously include around 20\% of LLM participants. 
In Table~\ref{tab:llm-wm-detection} we further report the AUROC for LLM-WM models broken down by LLM model backbone and Centaur.

\subsection{Discussion}

Our findings in the second simulation reveal that LLM participants that are either explicitly instructed to mimic working memory effects or are being finetuned on data from human psychology experiments qualitatively reproduce serial position and memory load effects on our working memory task.
Nevertheless, given access to a set of human participants, it is often possible to detect their anomalous quantitative working memory profiles, albeit imperfectly.
The cognitive anomaly detection approach we outline can easily be adapted to other cognitive domains with established cognitive effects.
It simply requires specifying a domain-specific Bayesian model of the cognitive mechanism that is being probed and access to a dataset of human participants.

\paragraph{Why not score using the \textit{joint} predictive density?}
On a technical level, our approach can likely be further improved by modeling the covariance of the random effects rather than assuming mutual independence in the prior.
We inherit the independence assumption from using the modeling library Bambi \citep{capretto_bambi_2022} within which it is, to the best of our knowledge, currently not easy to specify a prior on the full covariance matrix of the random effects.
The independence assumption of the random effects might also explain why we empirically find that using the \textit{pointwise} predictive density rather than the \textit{joint} predictive density works better for anomaly detection.
In principle, the latter should be more powerful since it takes into account the statistical dependencies between trials of the same participant.
However, due to the independence assumption on the random effects, participants with jointly abnormal random effects are not penalized if their marginals are within reasonable range.
At the same time the joint log predictive density rewards participants which behave consistently across all trials. %
As a result, the joint log predictive density can be comparably high for LLM participants, rendering anomaly detection based on it more difficult.

\paragraph{Why not compare reaction times?}
One might justifiably wonder why we did not attempt to distinguish LLM participants from online participants based on their reaction times.
A similar analysis could be applied to this analogous case and might likewise reveal differences between online participants and LLM participants.
For practical purposes it is certainly advisable to use reaction times.
However, LLM response times can vary significantly depending on the particular choices for model provider, model size, model type (e.g. reasoning vs. non-reasoning) and model harness.
As a result, we find it difficult to establish a robust comparison that generalizes beyond the particular choices we could make.
Given that inference-times for fixed capabilities have been steadily decreasing over the past years \citep{gundlach_price_2025}, it is conceivable that LLM participants instructed to behave humanlike will eventually be able to artificially manipulate their reaction times, e.g. via \texttt{sleep()} tool calls, to emulate human response times.

\section{General discussion}

Foundation models are becoming increasingly capable and can often overcome simple challenges such as logic puzzles or common knowledge questions designed to detect non-human participants in online behavioral experiments.
Our findings show that probing human cognitive constraints is a viable approach to detecting LLM participants and can possibly improve the validity of online behavioral research going forward.
Serial position and set size effects in working memory tasks are just one example of probing uniquely human cognitive constraints.
The cognitive science literature offers a rich resource for developing diverse paradigms for detecting non-human participants.
In practice, we will likely need to combine multiple measures for robust detection and ensure detection tasks are orthogonal to the research question to avoid introducing bias through imperfect detection.

More generally, the challenge of detecting humanness extends beyond online behavioral experiments.
With the rapid appearance of capable agents, humanness in online interactions can no longer be verified based on coherent behavior such as longform textual responses.
This poses a broader challenge for society since many of our institutions were built on the premise of being able to verify humanness through such means.
Cognitive science with its rich tradition of characterizing human behavior may play an important part in dealing with this challenge.

\paragraph{How can this be practically used?}
Our specific instantiation of the working memory task likely only remains effective for detection of LLM participants as long as human data from that instantiation is not used to adapt LLMs response patterns.
Luckily there are many ways to modulate human error patterns on this task, e.g. by varying the stimulus content, stimulus timing or the probe type.
However, it will be necessary to recollect human data on such variations, and ensure the data is not made publicly available to remain useful for anomaly detection. 

\paragraph{Will LLMs be able to imitate humans more faithfully as they become more capable?}
As LLM models are scaled, their capability to follow specific instructions and their breadth of knowledge has increased \citep{kaplan_scaling_2020, wei_emergent_2022}.
It is therefore conceivable that future model generations become more difficult to distinguish from humans when carefully instructed to imitate human cognitive constraints.
However, as outlined in the beginning, genuine cognitive constraints such as limited working memory are orthogonal to optimization pressures towards improving general model performance, e.g. on long-context reasoning or programming tasks.
In the absence of an inherent pressure to genuinely develop such constraints and without training on large quantities of data from cognitive science experiments, there is hope that detection based on the absence of human cognitive constraints remains viable.
\vspace{-1em}
\paragraph{What if LLMs are specifically trained to emulate human response patterns?}
Attempts at developing foundation models of cognition are ongoing \citep{binz_foundation_2025, bowers_misuse_2025, xie_centaur_2025}.
The fact that detection of Centaur participants was still largely possible suggests that the full promise of cognitive foundation models that generally predict human response patterns has yet to be realized despite our task being relatively standard and established in the literature.
However, as larger datasets from human psychology experiments are being collected and used to train cognitive foundation models, it will likely become more difficult to distinguish such models from human participants.
Moreover, using human cognitive constraints to detect LLMs online might inadvertently create an incentive towards developing such models.
If LLMs end up succeeding in reproducing human behavior on the breadth of tasks studied in cognitive science, this would imply that a good model of human behavior has been created, a major scientific achievement by itself.

\section{Code \& data availability}

Code for running the working memory online experiment as well as the LLM detection pipeline is publicly available at \textcolor{blue}{\url{https://github.com/smonsays/llm-humanness}}.
To mitigate the leakage of our data into training datasets, we only provide access upon reasonable request.

\section{Acknowledgments}

We would like to thank Suyog Chandramouli and Yassir Akram as well as the members of the Lake lab for fruitful discussions and helpful feedback on the project.
Google provided cloud compute credits for evaluating Gemini models.
Simon Schug was supported by Postdoc.Mobility grant P500PT\_225369 from the Swiss National Science Foundation.

\printbibliography

\end{document}